\documentclass[10pt,twocolumn,letterpaper]{article}

 \usepackage[pagenumbers]{cvpr} 
\usepackage{multirow}
\usepackage{xcolor,colortbl}
\usepackage{enumitem}
\definecolor{LightCyan}{rgb}{0.88,1,1}
\definecolor{yellow}{rgb}{0.91, 0.84, 0.42}
\definecolor{bubblegum}{rgb}{0.99, 0.76, 0.8}
\usepackage[font=small]{caption}
\usepackage{subcaption}

\definecolor{cvprblue}{rgb}{0.21,0.49,0.74}
\usepackage[pagebackref,breaklinks,colorlinks,citecolor=cvprblue]{hyperref}

\def\confName{CVPR}
\def\confYear{2024}

\title{Collecting Consistently High Quality Object Tracks with Minimal Human Involvement by Using Self-Supervised Learning to Detect Tracker Errors}

\author{Samreen Anjum$^{1,2}$ \quad Suyog Jain$^{3}$ \quad Danna Gurari$^{1,2}$ \vspace{0.3em} \\
{\normalsize $^1$University of Colorado Boulder} \quad
{\normalsize $^2$The University of Texas at Austin} \quad \\
{\normalsize $^3$FAIR, Meta}
}

\begin{document}
\maketitle

\begin{abstract}
We propose a hybrid framework for consistently producing high-quality object tracks by combining an automated object tracker with little human input. The key idea is to tailor a module for each dataset to intelligently decide when an object tracker is failing and so humans should be brought in to re-localize an object for continued tracking.  Our approach leverages self-supervised learning on unlabeled videos to learn a tailored representation for a target object that is then used to actively monitor its tracked region and decide when the tracker fails.  Since labeled data is not needed, our approach can be applied to novel object categories.  Experiments on three datasets demonstrate our method outperforms existing approaches, especially for small, fast moving, or occluded objects.  
\end{abstract}

\section{Introduction}
\label{sec:intro}

Object tracking entails identifying the trajectory of every object in a video. Applications include security surveillance~\cite{elhoseny2020multi}, traffic monitoring~\cite{jimenez2022multi}, and animal behavior analysis~\cite{patman2018biosense}. A practical challenge is that modern trackers often fail to produce \emph{consistently high quality} results, especially for novel object categories. In response, hybrid object tracking solutions have been proposed to combine automated methods with a little human input to curate object trajectories.  Such solutions can be valuable both for generating large-scale training datasets\footnote{Based on prior work's estimate that it could take 5.25 \cite{manen2017pathtrack} to 30 seconds \cite{su2012crowdsourcing} to annotate one bounding box, it would take at least 427 hours to create a standard tracking dataset like MOT16 \cite{milan2016mot16}, which has 14 videos.} as well as for providing high-quality tracks to end users.  Our work addresses two limitations of existing hybrid object tracking approaches. 

The first limitation of hybrid object tracking solutions stems from \emph{how human input is solicited}.  Existing approaches either lack intelligence or are predicated on access to human-annotated datasets.  For example, a long-standing, popular method~\cite{vondrick2013efficiently} is a na\"ive approach that uniformly samples frames in a video to annotate and so ignores the underlying visual content.  A recently introduced approach~\cite{kuznetsova2021efficient} instead intelligently selects frames that need to be annotated by using a trained model, however that model was developed using supervised learning and so is predicated on access to many human-annotated videos.  This requirement makes it challenging to generalize and deploy such a solution on videos that are different from those in the training sets.  For example, a model trained on videos showing people and cars would embed assumptions about visual representations that may not generalize well to videos showing animals or biological cells. 

The second key limitation of hybrid object tracking solutions is \emph{how they track objects}.  Existing approaches either lack intelligence or are predicated on access to human-annotated dataset showing the target object.  For example, approaches adopt geometric-based interpolation schemes such as linear interpolation \cite{vondrick2013efficiently} or customize the automated tracker for the target videos via supervised learning~\cite{kuznetsova2021efficient}. 

We propose a hybrid object tracking framework that addresses the aforementioned limitations of existing approaches.  It intelligently selects frames for humans to annotate without requiring access to human-annotated datasets showing the target object.  The key idea is to perform \textit{self-supervised representation learning} a priori to learn feature representations from a set of unlabeled videos that are tailored to the objects in those videos.  Using this model, the framework can then assess when a predicted object location seems to be mismatched in appearance from the object a tracker is trying to follow over time. We intentionally designed our framework to be agnostic to the underlying automated tracker in order to benefit from the ongoing progress around improving automated methods. The intention is to support integrating any off-the-shelf automated single object tracker (irrespective of its design, whether supervised, unsupervised, or self-supervised) into our framework without any additional customization, such as model retraining. 

The key contributions of this work are: (1) providing the first hybrid (i.e., human-machine collaboration) tracking framework for creating consistently high-quality tracking annotations that can be used with any automated tracker \emph{as is} (i.e., no extra training required) by leveraging self-supervised learning and track novel object categories, (2) experiments showing the versatility of our framework for use with different tracking algorithms on three datasets and that it outperforms all four of today's top-performing methods,  and (3) fine-grained analyses revealing that the advantage of our framework over prior work is in tracking small, fast moving, or highly occluded objects.

\section{Related Work}

\paragraph{Human annotation tools for object tracking.}
Modern object tracking annotation tools aim to limit human involvement through hybrid human-algorithm collaborations. Some methods require humans to annotate every object in every frame of the video.  For instance, PathTrack \cite{manen2017pathtrack} has users trace each object with a mouse pointer through the entire video and then converts those paths into bounding box tracks.  Other methods have humans annotate only a subset of frames and then employ algorithms to interpolate tracks for the remaining video frames~\cite{vondrick2013efficiently, yuen2009labelme, kuznetsova2021efficient}.  Our work contributes a new approach to this latter set of methods, called \emph{frame selection} methods.  Our experiments show the advantage of our approach over the prior work that annotates every object in every frame of the video (i.e., PathTrack~\cite{manen2017pathtrack}), as well as other frame selection methods ~\cite{vondrick2013efficiently, yuen2009labelme, kuznetsova2021efficient}.

\vspace{-0.75em}
\paragraph{Frame selection methods.} 
Different types of intelligence are used to decide which frames humans should annotate.  Some methods ignore the underlying visual content.  One popular example is uniform sampling, which emerged as a preferred choice after experiments showed that annotating a fixed set of frames sampled uniformly across a video leads to better results than having users select which frames to annotate~\cite{kuznetsova2021efficient,vondrick2013efficiently}.  Other methods consider the visual content.  Examples include ScribbleBox \cite{chen2020scribblebox}\footnote{This approach approximates low quality bounding boxes, which the authors claim are acceptable for the target downstream task.} and today's state-of-the-art approach~\cite{kuznetsova2021efficient}, which is a prediction system trained through supervised learning on both the underlying image content and an automated tracker's predicted object trajectory.  Our experiments demonstrate an advantage of our work over the existing methods that are both based on visual content~\cite{kuznetsova2021efficient} and not based on visual content (i.e., uniform sampling), achieving comparable or better results with less human involvement.  

Another related line of work selects optimal samples (target region) to train or update object tracking models~\cite{xin2019real, naiel2017online, liu2022reliable}.  Unlike our work, prior work uses static, one-size-fits-all measures to find the optimal frames that will improve a \emph{specific algorithm} to design a single \emph{fully-automated} model. Our approach, in contrast, leverages self-supervised learning to dynamically tailor features to novel object categories before selecting optimal frames for human intervention and is used at inference \emph{after deployment of any algorithm} for collecting consistently high-quality results.

\vspace{-0.75em}
\paragraph{Active learning.} 
A related approach to frame selection methods is to identify an optimal set of frames to annotate, when given a human annotation budget, for  training an object tracking algorithm~\cite{vondrick2011video}.  Such methods differ though because they focus on minimizing human involvement for developing algorithms rather than for collecting results at inference time after algorithms are deployed. %

\vspace{-0.75em}
\paragraph{Self-supervised representation learning in videos.}
Generally, self-supervised learning (SSL) methods aim to learn representations that are invariant to external factors such as illumination changes, scale transformations, or translations.  SSL methods are often studied in the context of downstream tasks. One important factor influencing the benefit of a learned representation is the dataset source used to train SSL models.  For example, images from ImageNet \cite{deng2009imagenet} typically contain single objects and so are valuable for learning image-level representations \cite{ramanathan2015learning, jayaraman2016slow}.  In contrast, more recent work \cite{gao2016object, yan2021learning, xie2021unsupervised} has shown that more powerful region-level or object-level representations are possible when learning from more complex images with more than one object \cite{xie2021unsupervised}. Such object-level representations have been shown to be beneficial for both the tasks of object detection \cite{yan2021learning} and image classification \cite{gao2016object}. 

Within the object tracking domain, SSL methods have been employed for fully-automated solutions for both single object tracking~\cite{vondrick2018tracking, yuan2020self, li2023self} and multiple object tracking \cite{bastani2021self, kim2023ssl,dev2023tracking}.  However, focusing on specific tracking algorithms is an orthogonal research direction, as our framework can be deployed with any tracking algorithm to permit the best option possible for a specific application.  While prior work uses 
measures to 
improve a \emph{specific tracking algorithm}, our work instead leverages learning through self-supervision to tailor the selection of optimal frames for human intervention for novel object categories to collect consistently high-quality results at inference time \emph{after deployment of any tracking algorithm}. Experiments show our framework's benefit of producing higher quality results than fully-automated trackers alone for state-of-the-art trackers.  






\begin{figure*}[t!]
    \centering
    \includegraphics[width = 1.0\linewidth]{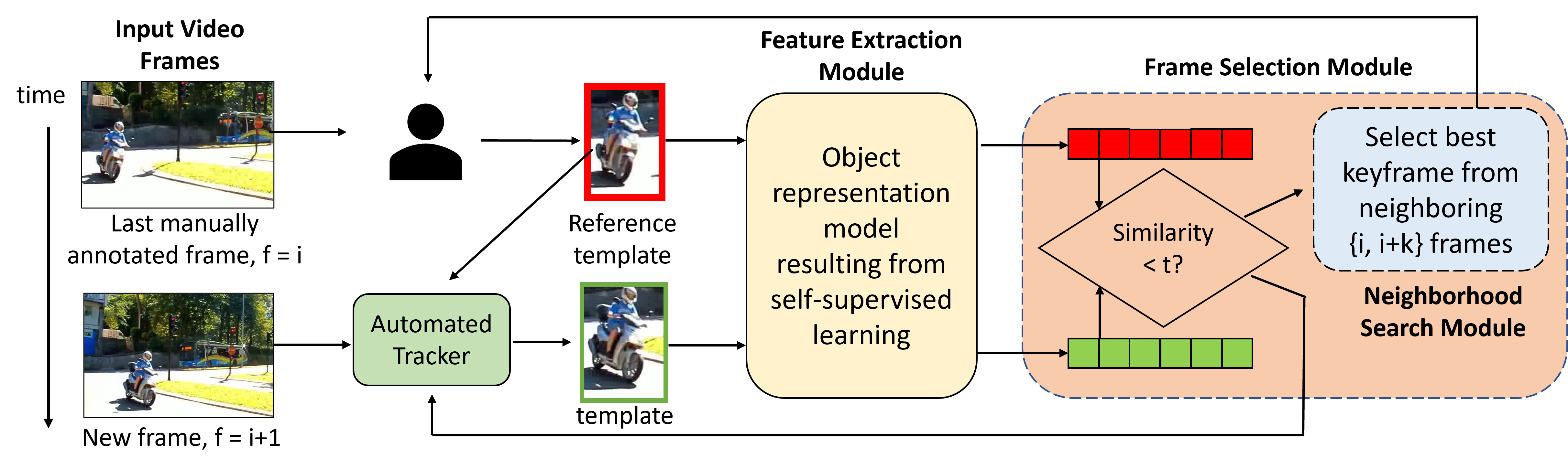}
     \vspace{-2em}
     \caption{Overview of our hybrid object tracking framework. The red bounding box represents a human drawn one and the green box represents an automated tracker's prediction. For each frame, the online frame selection module uses the self-supervised object representation model (trained offline) to compute features for the last reference template and automated tracker's template. If the feature similarity is below a threshold, a frame is selected for manual annotation using our neighborhood search module. Otherwise, the automated tracker continues on successive frames. } 
    \label{fig:approach}
\end{figure*}








\section{Method}
We now introduce our hybrid object tracking framework that combines automated tracking with minimal human involvement to generate high-quality tracks.  We call this framework \emph{SSLTrack} in order to capture its novel use of self-supervised learning (SSL) to limit human involvement.

\subsection{Overview}
We design our framework to support tracking both a single object and multiple objects.  An overview is shown in Figure~\ref{fig:approach}.  It centers on tracking one object at a time, because prior work \cite{vondrick2013efficiently, anjum2021crowdmot} showed that humans are better at annotating one object at a time across the entire video than multiple objects per frame.   Tracking is initiated by soliciting human annotations to localize each object of interest in the first frame each appears.  We refer to each human-annotated object as a \emph{reference template}. For each object, a single object tracking algorithm then uses the reference template to predict its location in the subsequent video frame.  Next, a \emph{feature extraction module} generates compact feature representations of both the human-annotated object and the object in its subsequently predicted location.  Then, a \emph{frame selection module} assesses whether the tracking algorithm may be failing by comparing the similarity of these representations.  When a high similarity is observed, the framework continues in a fully-automated fashion by repeating the process using the object's previously predicted location to predict its next location.  When a low similarity score is observed, the tracking algorithm may be failing and so a human is solicited to localize the object of interest.  The newly annotated region is then used as the object's new \emph{reference template} for successive analysis of the object.  This process is repeated for every tracked object until the end of a video.  

Two key components enable our framework's novelty of minimally involving humans for object tracking.  First, we developed a model for generating the compact representation of localized objects by using \textit{self-supervised learning} a priori (i.e., offline).  As a result, we do not rely on humans to label training datasets in order to represent objects to be tracked.  The second key component is our \textit{online} step of intelligently deciding when to solicit human input by predicting when the automated tracker is likely failing.  We offer a detailed description of each of these components below.

\subsection{Learning an object representation model}
We considered two key motivating factors when designing our model for compactly representing objects.  First, we wanted to be able to deploy it in the wild for novel object categories without curating human-annotated datasets to train models.  Additionally, the generated representations had to be effective for assessing whether two cropped regions from two video frames show the same object at two different time points (i.e., successful object tracking).  


For our offline process of learning this model, we leverage the self-supervised learning approach described in \cite{wei2021aligning}\footnote{We explored several approaches for object representation, which will be presented in Section 4.2, and found the object-proposal based approach is best suited to capture object properties for our purpose.}.  While originally the approach was used for object detection, we explore its benefit as well for object tracking. Specifically, we train a model to learn object representation models that are tailored to our target videos with novel object categories.  To do so, we first extract hundreds of object-like regions from every frame of the unlabeled videos using the selective search approach described in \cite{wei2021aligning}\footnote{We use the SoCo package \cite{wei2021aligning} with default parameters and two modifications (i.e., aspect ratio and size of filtered proposals). We retain object proposals of any size since many objects in our experimental datasets are small (e.g., mean relative size of an object compared to the input image is 4.45$\%$). Also, due to the long processing times of the selective search method, we only extract proposals for the first 100 frames of every video.}.  By generating region proposals for every frame, we aim to capture a diversity of any relevant object's potential appearance over time.  We then feed these object proposals to the self-supervised learning algorithm, SimCLR, which is a state-of-the-art contrastive-based representation learning method that was reported to be one of the top performing methods in a recent benchmark analysis \cite{jaiswal2020survey}\footnote{We use the lightly package \cite{lightly} implementation of the SimCLR model with default parameters as listed in their documentation.}.

\subsection{Deciding when to solicit human input}
If the feature similarity between the last reference template at frame $I_i$ and the template predicted by the tracker at a new frame $I_{i+1}$  is below a certain threshold, we consider this as a cue that the tracker may have failed and begin the frame selection process.  While we initially selected the frame $I_{i+1}$ for human annotation, we found that this can lead to a sensitive, inefficient outcome where multiple consecutive frames are selected for human annotation. For example, this can occur when the target object barely moves in consecutive frames but there are significant changes in the background.  Inspired by the Non-Maximum Suppression technique (NMS) \cite{girshick2014rich} in object detection, we instead propose a neighborhood search (NS) approach to have a human annotate only one keyframe from a set of neighboring frames. When a candidate keyframe for human annotation is identified (i.e $I_{i+1}$), we continue the automated tracking process for an additional $k$ frames forward. We then compare the similarity of the tracked object at each of the k frames with the reference template at $I_i$, and select the frame with the lowest score for human annotation. 

\subsection{Implementation details}
There are several implementation choices for our framework. First, in choosing an automated single object tracker, we consider two trackers to motivate the potential of our framework to generalize for any off-the-shelf object tracker: OSTrack  \cite{ye2022joint} and Stark \cite{yan2021learning} with their default parameters \cite{ostrackgithub, starkgithub}, which were the best publicly-available algorithms for the Visual Object Tracking Challenge \cite{VOT_TPAMI} 2022 and 2021 respectively\footnote{To motivate our framework's potential benefit in a multi-object tracking scenario, we analyze a single object tracker's ability to maintain an object's identity across a video by measuring the number of identity switches, meaning when a bounding box of an object covers a ground truth bounding box of a different object.  Errors were small for three datasets, ranging from 0.1\% to 1.8\%.  Details are provided in the Supplementary Materials.}.  For assessing the similarity between the representations of an object's reference template and its future predicted location, we chose the cosine distance.  For the \emph{Frame Selection Module}, values must be chosen for both the similarity threshold, $t$, and the size of the neighborhood, $k$. Better tracking performance is expected by increasing the similarity threshold and reducing the neighborhood size, while less human involvement is expected by reversing these trends.  We explore this trade-off in our experiments with two different combinations of values for the similarity threshold and neighborhood size. Finally, we include an optional offline, post-processing step, where bounding boxes are linearly interpolated between keyframes that are temporally close to each other.  This is particularly valuable for tracks with heavy occlusions where the object tracker may produce unreliable results \cite{kuznetsova2021efficient}.

\section{Experiments}
\label{sec:experiments}
We now investigate the performance of our approach.  We conduct all experiments with RTX8000 GPUs. 

\vspace{-0.75em}\paragraph{Datasets.}
We conduct our experiments with three mainstream object tracking datasets.

First, we leverage the recently published \emph{GMOT-40} \cite{bai2021gmot}. It contains 40 videos showing 10 different object categories, with a mean video length of 240 frames and all 40 videos showing multiple objects. The mean relative size of objects is 0.63\% when compared to the image size.  This dataset is challenging due to high object density (i.e., mean of 26 per frame with up to 100 objects in a frame) with often similar looking, small objects and many occlusions. 

We also leverage the popular \emph{ImageNet VID} \cite{russakovsky2015imagenet}. We use the validation split of this dataset that consists of 555 videos showing 30 different types of objects, with a mean video length of 317 frames and 60\% of the videos showing single objects. The mean relative size of objects is 11.2\% when compared to the image size. 

Next, we leverage the popular \textit{MOT15} \cite{leal2015motchallenge}. We use the training split that consists of 11 videos showing pedestrians, with a mean video length of 500 frames and all 11 videos showing multiple objects. The mean relative size of objects is 2.7\% when compared to the image size. This dataset offers several challenging tracking scenarios including high object density (i.e., mean of 45 per video with some frames showing more than 100 objects) and many occlusions.

\vspace{-0.75em}\paragraph{Human annotation.}
We use the ground truth bounding box annotations in each dataset to represent humans.

\subsection{Overall Performance}
We first compare our approach to the following methods:

\begin{itemize}[itemsep=0.1em,leftmargin=*]
\item \textit{State-of-the-art, human-in-the-loop approach}~\cite{kuznetsova2021efficient}: unlike our approach, it requires supervised training of models on human-labeled object tracking datasets to design a customized automated tracking module and frame selection module. Since the code is not publicly-available, we can only report its performance on the setting published in \cite{kuznetsova2021efficient} on the MOT15 and ImageNet VID datasets.

\item \textit{PathTrack}~\cite{manen2017pathtrack}: it converts human tracings of every object in every video frame into bounding box tracks.  Since the code is not publicly-available, we can only report its performance on the setting published in \cite{manen2017pathtrack}.

\item \textit{Uniform sampling / VATIC}~\cite{vondrick2013efficiently}: it uses a constant interval to sample video frames for humans to annotate.  We use both OSTrack \cite{ye2022joint} and Stark \cite{yan2021learning} as the automated trackers and choose a frame sampling interval such that the mean number of manually annotated boxes per object matches those used by our proposed approach. We also report performance using the setting published in \cite{manen2017pathtrack}.

\item \textit{LabelMe}~\cite{yuen2009labelme}: this framework relies on users to select keyframes. For consistency, we again report its performance on the setting published in \cite{manen2017pathtrack}. 

\item \textit{Fully-automated approach}: our framework's single object trackers, OSTrack and Stark, which reflects no human intervention except for the first frame each object appears. 
\end{itemize} 

To explore the versatility of our approach, we show results in the main paper for two combinations of values for the similarity threshold, $t$, and the neighborhood size, $k$, of the \emph{Frame Selection Module}: $\{t, k\}$ = $\{0.95, 20\}$, $\{0.97, 10\}$\footnote{We explore a total of three combinations but, due to space constraints, report results with a discussion in the Supplementary Materials.}. These signify being less sensitive (i.e., $\{t, k\} = \{0.95,20\}$) to more sensitive (i.e., $\{t, k\} = \{0.97, 10\}$) to changes in the target object's appearance.  

\vspace{-0.75em}\paragraph{Evaluation metrics.}
 We use the standard metrics employed in the single-object tracking (SOT) and multi-object tracking (MOT) communities: i.e., recall and Multiple Object Tracking Accuracy (MOTA) \cite{bernardin2008evaluating} respectively.  \textit{Recall@0.7}, as introduced in \cite{kuznetsova2021efficient}, measures the object detection performance by computing the fraction of bounding boxes that have an intersection-over-union score with the ground truth higher than 0.7.  \textit{MOTA} \cite{bernardin2008evaluating}~\footnote{We use the evaluation package provided by \textit{py-motmetrics} \cite{py-motmetrics}, and report other MOT metrics (CLEAR \cite{bernardin2008evaluating}) in the Supplementary Materials.} represents object coverage across the video as follows:
    \[MOTA = 1 - \frac{\sum_{t}(FN_t+FP_t+IDSW_t))}{\sum_{t}GT_t}\]
\noindent
where FP is the number of predicted boxes not matched by any ground truth (GT) box, FN is the number of GT boxes not matched by any predicted box and IDSW is the number of identity switches, i.e., a bounding box matching a GT box from a different track than the target track.  MOTA scores can range between negative values and 100, with higher values indicating better performance.  

We also compute annotation time, as defined in \cite{manen2017pathtrack}: $t_{track} = \lambda t_{watch} + t_{box}.N_{box} $ where, $t_{watch}$ is the time taken to watch a track in the entire video, $t_{box}$ is the average annotation time per box (i.e., 5.2s), and $N_{box}$ is the total number of boxes drawn by the human per track.

\begin{figure}[t!]
    \centering \includegraphics[width=0.9\linewidth]{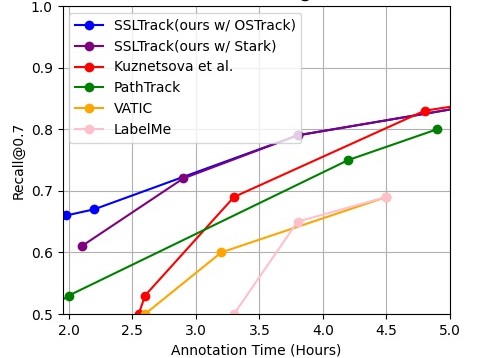}
    \caption{Performance comparison of our method against other state-of-the-art methods on MOT15 dataset. Our method using both automated trackers, OSTrack and Stark, outperforms all other methods and achieves higher accuracy at a lower annotation rate.}
    \label{fig:hybrid_comparison}
\end{figure}

\vspace{-0.75em}\paragraph{Overall results.}
Results are in Figure~\ref{fig:hybrid_comparison} and Tables~\ref{tab:cost}-\ref{tab:mota_scores_ILSRVC}. 

In the setting where comparison to all prior work is possible\footnote{Recall, we can only support comparison based on results reported in several baselines' publications since there code is not publicly available.}, our approach achieves the highest recall with less human effort.  This is true when assessing performance with both tested trackers (Figure~\ref{fig:hybrid_comparison}).  The advantage of our approach is even greater when less human time is available. 

\begin{table*}[t]
\scriptsize
\centering
\begin{tabular}{|c|c|ccc|cc|ccc|}\hline
\multirow{2}{*}{\textbf{FS Method}} & \multicolumn{1}{c|}{\multirow{2}{*}{\textbf{Automated Tracker}}} & \multicolumn{3}{c|}{\textbf{Training FS method on ImageNet VID}} & \multicolumn{2}{c|}{\textbf{Reported scores on ImageNet VID}} & \multicolumn{3}{c|}{\textbf{Annotating a large dataset of 1M objects}} \\ \cline{3-10} 
 & \multicolumn{1}{c|}{} & \textbf{\#boxes} & \textbf{Time (hours)} & \textbf{Cost (\$)} & \textbf{\#boxes/obj} & \textbf{Recall} & \textbf{\#boxes} & \textbf{Time (hours)} & \textbf{Cost (\$)} \\\hline
K et al. \cite{kuznetsova2021efficient} & Customized & 1M & 1,500 & 22,500 & 4.9 & 0.81 & 4.9M & 7,146 & 107,190 \\
Uniform \cite{vondrick2013efficiently} & OSTrack \cite{ye2022joint} & 0 & 0 & 0 & 5.6 & 0.83 & 5.6M & 8,167 & 122,505 \\
\textbf{Ours} & \textbf{OSTrack\cite{ye2022joint}} & \textbf{0} & \textbf{0} & \textbf{0} & \textbf{4.5} & \textbf{0.83} & \textbf{4.5M} & \textbf{6,563} & \textbf{98,445}\\\hline
\end{tabular}
\caption{Comparison of our method to the state-of-the-art methods for tracking a large dataset containing 1M objects matching the properties of ImageNet VID. 
Compared to \cite{kuznetsova2021efficient}, our method solicits less human input per object, saves money, and yields \emph{higher performance}.
}
    \label{tab:cost}
\end{table*}

In the other setting where comparison to the state-of-the-art hybrid tracker~\cite{kuznetsova2021efficient} is also possible, our method achieves a higher recall value of 0.84 and 0.81 using off-the-shelf trackers OSTrack\cite{ye2022joint} Stark\cite{yan2021learning} respectively, with considerably less human involvement  (row 1 in Table \ref{tab:cost} and row 8 in Table \ref{tab:mota_scores_ILSRVC}). We attribute this to our approach's reliance on \emph{self-supervised} learning rather than \emph{supervised} learning on human-labeled video datasets like prior work~\cite{kuznetsova2021efficient}.  To further illustrate the difference in human involvement, consider ImageNet VID alone which consists of over 1M video frames.  While there would be no human or monetary cost for our approach, it would cost prior work approximately 1,500 annotation hours and \$22,500 to train the approach to perform frame selection (Table \ref{tab:cost})\footnote{We assume each frame has one bounding box to annotate, it takes 5.25 seconds to annotate one box, and costs \$0.02 to annotate each bounding box to achieve a wage of \$15/hour.}. As another example, consider annotating a video collection with 1M objects.  Not only would our method save money (e.g., training costs and \$8,745 at inference time) due to less human input per object (e.g., 4.5 vs. 4.9 boxes/obj), but it also would yield higher performance (e.g., 0.83 vs. 0.81 recall).

\newcommand{\lc}{\cellcolor{LightCyan}}
\newcommand{\gum}{\cellcolor{bubblegum!50}}

\begin{table}[t!]
\scriptsize
\centering
\begin{tabular}{cl|ccc|ccc}\hline
\multicolumn{1}{c}{\multirow{2}{*}{\textbf{Data}}} & {\multirow{2}{*}{\textbf{FS}}}   & 
\multicolumn{3}{c}{\begin{tabular}[c]{@{}c@{}}\textbf{Stark\cite{yan2021learning}}\end{tabular}} &
\multicolumn{3}{c}{\begin{tabular}[c]{@{}c@{}}\textbf{OSTrack\cite{ye2022joint}}\end{tabular}}
\\ \cline{3-4} \cline{5-8}
  &    &  \#b/obj & Recall & MOTA  &  \#b/obj & Recall & MOTA                            \\ \hline
\multirow{5}{*}{\rotatebox[origin=c]{90}{GMOT-40}} &     Auto & 1& 0.37 & 18.1 & 1 & 0.42 & 30.9\\ \cline{2-8}&
Uni      & 3.9  & 0.62& 60.4 & 3.9 &0.67 &71.1\\
& 
\lc Ours& \lc3.9   & \lc\textbf{0.63} &
\lc\textbf{62.1}   & \lc3.9 &  \lc\textbf{0.68} &   \lc\textbf{72.9}    \\ 
\cline{2-8}   & 
Uni   & 6.9  &0.70 & 72.3 & 6.9 & 0.74 & 81.3 \\ & 
\gum Ours  & \gum6.9   & \gum\textbf{0.71}& \gum\textbf{75.3}     & \gum6.9 & \gum\textbf{0.75}   & \gum\textbf{83.4}     \\ \hline
\multirow{5}{*}{\rotatebox[origin=c]{90}{ImageNetVID}}    & 
Auto & 1& 0.69 &74.3  & 1 & 0.74 & 78.4 \\ \cline{2-8}&
 Uni   & 5.6  & 0.80& 87.8   & 5.6 & 0.83 & 90.8
 \\  & 
 \lc Ours     & \lc5.6& \lc\textbf{0.81}& \lc\textbf{89.4}   & \lc5.6 & \lc\textbf{0.84} &  \lc\textbf{92.1} \\ 
 \cline{2-8}   &  
 Uni & 10.4  & 0.83& 91.6 & 10.2 & 0.86 & 94.1  \\  & 
 \gum Ours   & \gum 10.4&\gum \textbf{ 0.84}& \gum \textbf{93.1}   & \gum10.2 &  \gum\textbf{0.87}&  \gum\textbf{94.7}           
 \\ \hline
 \multirow{5}{*}{\rotatebox[origin=c]{90}{MOT15}}    & 
 Auto  & 1& 0.46 &41.9 &1 & 0.55 & 57.6\\ \cline{2-8}&
 Uni   & 2.84  & 0.61&  67.4   & 2.92& 0.67 & 77.5 
 \\  & 
 \lc Ours                    & \lc 2.84 & \lc\textbf{0.62}& \lc\textbf{68.2} & \lc 2.92 & \lc\textbf{0.69} & \lc \textbf{78.4}                
 \\ 

 \cline{2-8}   & 
 Uni  &  5.06 &  0.68& 78.0 & 5.03& 0.71& 82.8  \\  & 
 \gum Ours                    & \gum 5.06 & \gum \textbf{0.79}& \gum \textbf{89.1} & \gum 5.03& \gum \textbf{0.79}& \gum \textbf{89.4}    
 \\\hline
\end{tabular}
\caption{Performance of our frame selection (Ours), the fully-automated approach (Auto), and uniform frame sampling (Uni) (FS = Frame Selection Method, \#b/obj = mean number of boxes manually annotated per object track). The colors represent two different hyperparameters settings ($t$, $k$) used for selecting frames for manual annotation ($t$ = similarity score threshold, $k$ = number of neighboring frames considered in the neighborhood search approach. Cyan and pink represent the following $(t,k)$: $(0.95, 20)$ and $(0.97, 10)$,  respectively).
Our method outperforms both baselines across all three datasets and two different automated trackers. }
\label{tab:mota_scores_ILSRVC}
\end{table}

For the remaining analysis, only comparison to uniform sampling (i.e., VATIC) and fully-automated methods is possible.  As expected, fully-automated methods perform the worst on all datasets\footnote{Our findings also reveal that GMOT-40 and MOT15 are more challenging datasets than ImageNet VID for our chosen automated tracker.}, which confirms the advantage of human-in-the-loop approaches for tracking objects.  Our approach of \emph{intelligently} soliciting human input consistently outperforms uniform sampling across both recall and MOTA metrics.  For example, we see in Table \ref{tab:mota_scores_ILSRVC}, when soliciting human input for just 5-7\% of frames per object, the annotation quality with our approach improves by 11, 3 and 1.5 MOTA percentage points on MOT15, GMOT-40 and ImageNet VID, respectively with Stark, and 7, 2, and 0.6 percentage points with OSTrack.  In more concrete terms, we observe that for ImageNet VID, our method with OSTrack generated a similar quality of object trajectories (i.e., MOTA=$\sim$91\%) using only 4.5 boxes per object whereas uniform sampling required 5.6 boxes per object (i.e., 24\% more annotation time). This translates to saving at least one box annotation per object. Given a dataset with 1M objects, manually annotating even one extra box per object with uniform baseline translates to an additional 1,500 annotation hours and \$22,500 compared to our method (Table \ref{tab:cost})\footnote{Assumes a rate of 5.25 secs/box and a minimum wage of \$15/hour.}. 

From visual inspection, the advantage of our hybrid approach is that it selects a frame for human annotation \emph{only} when the target object undergoes a considerable appearance change.  In contrast, uniform method selects frames at equal intervals, thereby neglecting to collect more when they would be useful and wasting annotation when there is a negligible change (e.g., when an object is not moving). 

\vspace{-0.75em}\paragraph{Analysis with respect to object attributes.}
We next conduct fine-grained analyses based on four object properties in the video: size, displacement, occlusion, and orientation. We combine the objects from all datasets for this analysis, resulting in a total of 3,823 objects\footnote{While we conduct this analysis for the both trackers, due to space constraints and similar trends across both trackers, we only report findings for the better-performing OSTrack here.}.  For size, we compute the relative size of the object compared to the image. For displacement, we compute the difference between the centroids of the object in consecutive frames and take an average across the entire track in the video. To compute the level of occlusion, we follow the approach used to annotate the MOT16 dataset \cite{milan2016mot16}. For changes in orientation, we compute the difference between centroids of an object in consecutive frames and analyze the difference independently along the X and Y axis.   

\begin{figure}[t!]
    \centering
    \includegraphics[width = 0.95\linewidth]{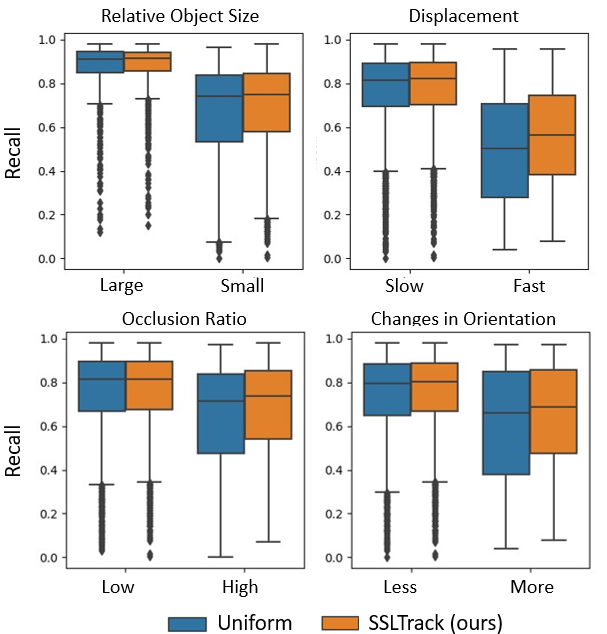}
    \caption{Tracking performance with respect to object attributes for all the objects in three datasets. Relative object size is the ratio of the object size with respect to the input image, displacement is the speed of the object, occlusion ratio is the level of occlusion, and changes in orientation is the number of times an object changes its orientation across the video. We see that our method outperforms the uniform baseline approach particularly on challenging scenarios when objects are smaller, moving fast, encounter occlusions, and change their orientation more in their trajectories.}
    \label{fig:obj_attributes}
\end{figure}

For each measure, we compute the mean of each attribute and then divide the objects into two categories based on whether they are below or above the mean value (Figure \ref{fig:obj_attributes}).  The mean relative size of the objects is 4.45\%. We label all objects with relative size less than the mean as ``Small" and greater as ``Large". The mean displacement is 13.73. We again label all objects below mean as ``Slow" and above average as ``Fast". The mean occlusion ratio is 0.31. Objects below the mean occlusion ratio are labeled as ``Low" and above are labeled as ``High". The mean number of changes in orientation of an object is 4.27. Objects below and above this value are labeled as ``Less" and ``More" respectively. 

Results are shown in Figure \ref{fig:obj_attributes}.  We observe that both frame selection methods find objects that are larger easiest to track while faster moving objects are the most difficult.  More generally, objects which are smaller, faster, more occluded or changing orientation more challenging to track.  Consequently, these are promising research challenges for future work.  We find that our method outperforms uniform sampling by a larger margin (i.e., $>$4\%) on these challenging types of objects, which suggests that our method is better suited for more challenging tracking scenarios. 

\begin{figure}[t!]
    \centering
    \begin{subfigure}[t]{0.45\textwidth}
        \centering
        \includegraphics[width=\textwidth]{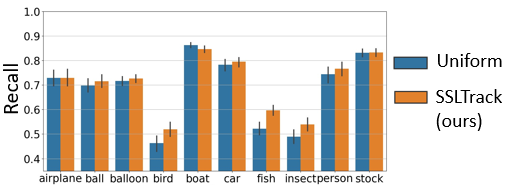}
        \caption{GMOT-40}
   \end{subfigure}
   ~
    \begin{subfigure}[t]{0.45\textwidth}
        \centering
       \includegraphics[width=\textwidth]{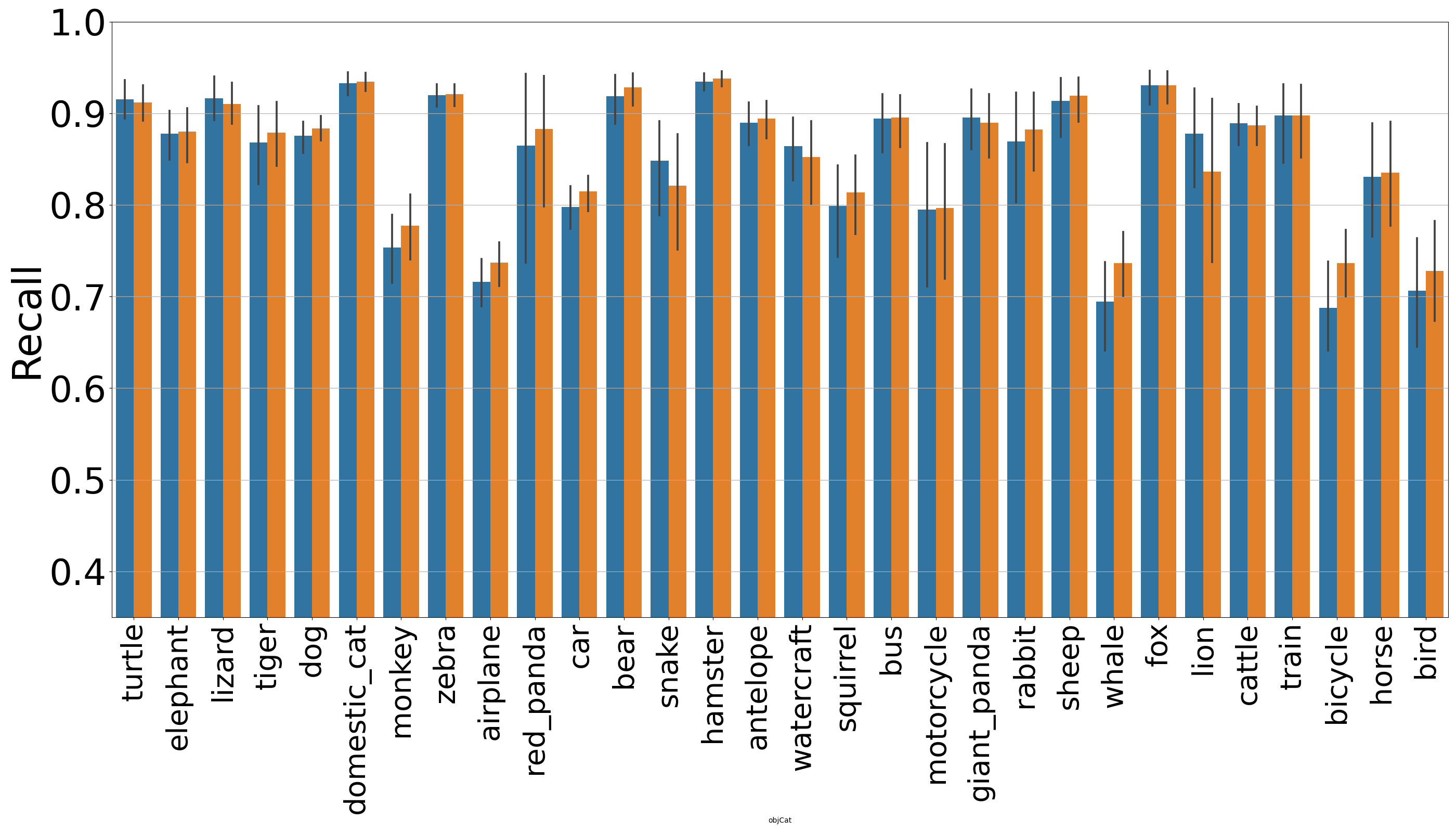}
        \caption{ImageNet VID}
    \end{subfigure}
    \caption{Performance comparison of our method against the uniform selection across different categories in both datasets. Our method outperforms the baseline on 8 out of 10 categories for GMOT-40 and 23 out of 30 categories for ImageNet VID.}
    \label{fig:category}
\end{figure}

\vspace{-0.75em}\paragraph{Category-specific analysis.}
We next evaluate with respect to each object category in GMOT-40 and ImageNet VID, comparing the uniform baseline with our approach\footnote{MOT15 only contains ``person" category so is not relevant for this analysis.  Due to space constraints and similar trends for both trackers, we only report findings for the better-performing OSTrack here.}. Altogether, we demonstrate performance of our approach across 36 unique novel categories. Results are shown in Figure \ref{fig:category}. Overall, our method outperforms the baseline in 8 out of 10 categories and 23 out of 30 categories in GMOT-40 and ImageNet VID, respectively.  

For GMOT-40, on videos showing fish, birds, and insects, our method scored 7.4, 5.6, and 5 percentage points (recall) better than the uniform baseline. We highlight this considerable improvement on these categories since these were the most challenging categories for both frame selection methods (i.e., ours and uniform) from the 10 categories.  These objects are challenging in part because they moved faster (above the mean displacement of 13.7), which as noted in the analysis about object attributes, is a scenario where our method outperforms the baseline. Prior work~\cite{bai2021gmot} also indicated that insect and bird were the most challenging categories in GMOT-40 for automated MOT trackers.

We observe similar trends for the ImageNet VID dataset between object attributes and categories. For example, our approach outperforms the baseline by over 5 and 4 percentage points (recall) for videos showing bicycles and whales respectively. These object categories are relatively smaller in size and are occluded more frequently. 

\vspace{-0.75em}\paragraph{Analysis with respect to automated trackers.} 
Overall, our framework outperforms all baselines when used with both automated trackers. This underscores its potential for use with any state-of-the-art trackers. We also observe that performance is better with OSTrack than with Stark, underscoring a benefit of leveraging the best tracker possible for a particular dataset (i.e., OSTrack outperforms Stark on all three datasets). For example, in Table \ref{tab:mota_scores_ILSRVC}, for the same annotation effort of 6.9 boxes per object, we see on GMOT-40 that our framework achieves a 4 point boost in tracking quality (i.e., 0.75 with OSTrack vs. 0.71 with Stark). On the other hand, with ImageNet VID, our method achieves the same recall rate of 0.84 at a reduced annotation effort of 5.6 with OSTrack vs. 10.4 with Stark. This suggests our method can be used with future, improved automated trackers that will, in turn, reduce human effort even more.  

\vspace{-0.75em}\paragraph{Analysis with respect to videos showing SOT vs. MOT.}
Of the 606 videos across all datasets, 54\% showed single objects while the rest contained multiple objects. We found our method performed comparable against uniform on videos showing single objects at a recall rate of 0.91 and performs better than uniform on videos showing multiple objects by 2 percentage points (0.71 vs. 0.69). This reaffirms our prior observation that the primary benefit of our approach is for more challenging MOT scenarios.  Yet, as shown, it still preserves high quality results that can be expected from today's baselines for easier tracking scenarios. 





\subsection{Analysis of Object Representation Models} 

\begin{figure}
    \centering    \includegraphics[width=0.95\linewidth]{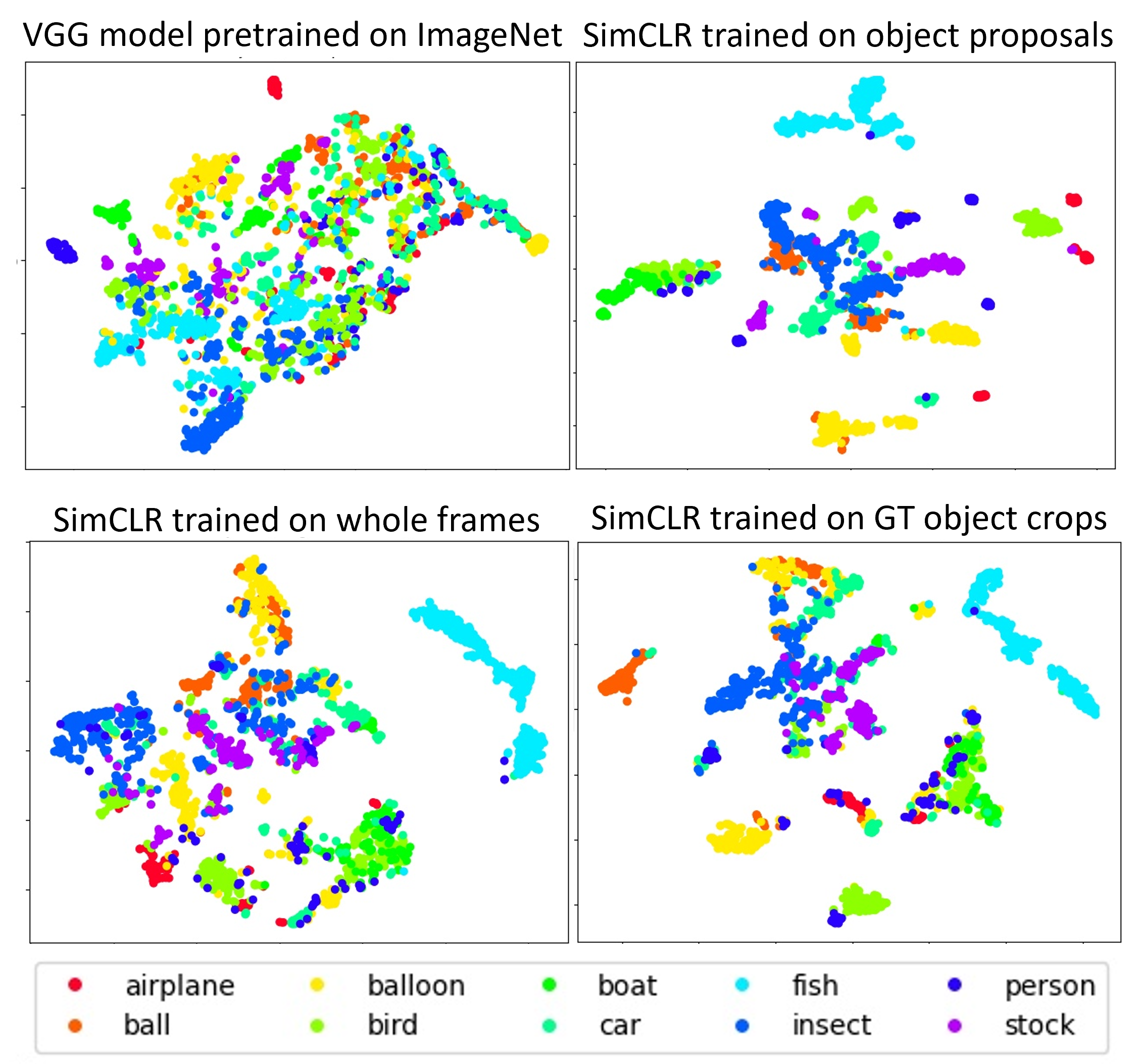}   
    \caption{t-SNE visualization of object embeddings from four different representation models. Each data point in each map represents the ground truth crop of a unique object in the GMOT dataset (total = 1,944). Embeddings are extracted using (a) off-the-shelf VGG-16 model pre-trained on ImageNet, (b) SimCLR model trained on object proposals extracted from GMOT-40 (our model), (c) SimCLR model trained on whole frames from GMOT-40, and (d) SimCLR model trained on all ground truth instance crops in the GMOT-40 dataset. Embeddings extracted using our model is better separated than the VGG model and SimCLR model trained on whole frames, while similar to the one trained on GT crops (e.g., balloon, bird). Best viewed in color.}
    \label{fig:embeddings}
\end{figure}

We now investigate the importance of our choice for an object representation model. To do so, we use t-SNE to visualize object embeddings extracted from our model and compare it against three different representation models.

For this analysis, we compare our SimCLR model trained on object proposals extracted from input frames in GMOT-40 videos against a VGG-16 model pre-trained on the ImageNet dataset, a SimCLR model trained on all video frames in GMOT-40, and a SimCLR model trained on the ground truth bounding box regions of all objects in GMOT-40. We use GMOT-40 dataset for this analysis as this dataset consists of multiple object categories and was not used to train the off-the-shelf VGG-16 model and so, allows us to understand the impact of using an off-the-shelf model. We also compare with a SimCLR model trained on whole frames to demonstrate why learning on whole frames does not capture the object-specific properties that are valuable for our goal of object tracking.  
We collect one ground truth bounding box region from each unique object in GMOT-40, resulting in a total of 1,944 instances. We then extract features for each of these instances using the four models.

Figure \ref{fig:embeddings} shows the resulting t-SNE visualizations.\footnote{More visualizations are provided in the Supplementary Materials.} We see that embeddings extracted using our model are better separated than those based on the off-the-shelf VGG-16 model and the model trained on whole frames, while similar to the one trained on GT crops. This reinforces our core observation that object-specific representation models are well-suited to represent objects for our tracking problem. 

\subsection{Ablation Study}
\begin{figure}[t!]
    \centering
    \includegraphics[width = 1.0\linewidth]{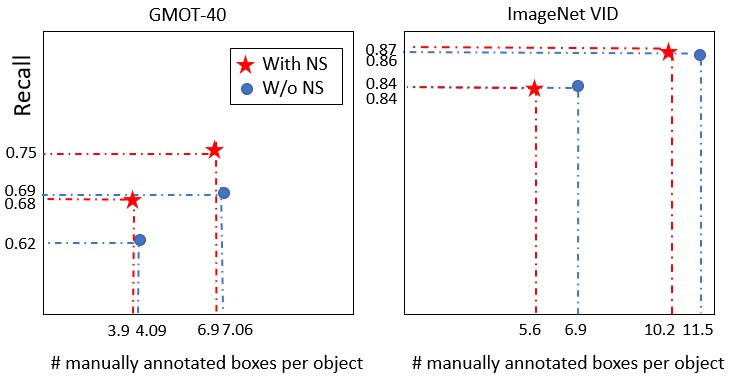}
    \caption{We compare our frame selection method with and without the integration of the neighborhood search (NS) approach. We see that with the NS method, we achieve a higher accuracy with a lower annotation effort in GMOT-40, and comparable accuracy with lower annotation effort in ImageNet VID.   }
    \label{fig:kfablation}
\end{figure}

We conduct an ablation study to demonstrate the importance of the neighborhood selection module by comparing our method with and without this module on two datasets\footnote{MOT15 results are provided in the Supplementary Materials.} with OSTrack. For the experiment with NS, we use the same results as reported in Table \ref{tab:mota_scores_ILSRVC}. For without NS, we experiment both datasets with $t=0.85$ and $0.90$ for comparable human effort. Results are shown in Figure \ref{fig:kfablation}.

Overall, the NS module enables reliance on less human annotation while achieving comparable (for ImageNet VID) or better (for GMOT-40) tracking performance. We suspect the performance boost for GMOT-40 is due in part because, when objects are very similar in appearance, the NS approach enables us to assign a higher threshold for feature similarity to more closely monitor the tracker's performance while lowering the risk of over-annotation by humans by preventing selection of many consecutive frames. Without the NS module, our approach is susceptible to over-selection of frames, as observed for both datasets.

\section{Conclusion}
We propose a hybrid object tracking framework that intelligently selects frames for human annotation using self-supervised representation learning.  Experiments demonstrate it outperforms existing approaches on three datasets.  



{
    \small
    \bibliographystyle{ieeenat_fullname}
    \bibliography{main}
}


\end{document}